\pdfoutput=1

\documentclass[11pt]{article}

\usepackage[preprint]{acl}

\usepackage{times}
\usepackage{latexsym}
\usepackage[ruled,linesnumbered]{algorithm2e}
\usepackage[T1]{fontenc}

\usepackage[utf8]{inputenc}

\usepackage{microtype}

\usepackage{inconsolata}
\usepackage{algorithmic}

\usepackage{graphicx}
\usepackage{multirow}
\usepackage[normalem]{ulem}
\usepackage{arydshln}
\usepackage[section]{placeins}
\usepackage{float}
\usepackage{placeins}
\usepackage{soul} 
\usepackage{needspace}
\usepackage{mwe} 
\usepackage{cuted}
\usepackage{caption}

\title{RLJP: Legal Judgment Prediction via First-Order Logic Rule-enhanced with Large Language Models}

{
\author{
 \textbf{Yue Zhang\textsuperscript{1}},
 \textbf{Zhiliang Tian\textsuperscript{1}},
 \textbf{Shicheng Zhou\textsuperscript{2}},
 \textbf{Haiyang Wang\textsuperscript{1}},
\\
 \textbf{Wenqing Hou\textsuperscript{1}},
 \textbf{Yuying Liu\textsuperscript{1}},
 \textbf{Xuechen Zhao\textsuperscript{3}},
 \textbf{Minlie Huang \textsuperscript{4}},
 \textbf{Ye Wang\textsuperscript{1}},
\\
 \textbf{Bin Zhou\textsuperscript{1}} 
\\
\\
\textsuperscript{1} \small{College of Computer Science and Technology, National University of Defense and Technology. } \\\small{ No.137 Yanwachi Street, Changsha, Hunan, 410073, P R.China,}\\
 \textsuperscript{2}\small{College of Electronic Engineering, National University of Defense Technology.}\\\small{ No. 460, Huangshan Road, Shushan District, Hefei. 230037, P. R. China,}\\
 \textsuperscript{3}\small{School of Data and Computer Science, Shandong Women's University. No.2399 Daxue Road, Ji'nan, Shandong, 250300, P R.China}\\
 \textsuperscript{4}\small{Institute for Artificial Intelligence, Tsinghua University.}\\ \small{ Dept. of Computer Science, Tsinghua University, Beijing 100084, China}
\\
 \small{
   \textbf{Correspondence:} \href{binzhou @nudt.edu.cn}{binzhou @nudt.edu.cn}   
 }
}}

\begin{document}
\maketitle
\begin{abstract}

Legal Judgment Prediction (LJP) is a pivotal task in legal AI. 
Existing semantic-enhanced LJP models integrate judicial precedents and legal knowledge for high performance. But they neglect legal reasoning logic, a critical component of legal judgments requiring rigorous logical analysis. 
Although some approaches utilize legal reasoning logic for high-quality predictions, their logic rigidity hinders adaptation to case-specific logical frameworks, particularly in complex cases that are lengthy and detailed. 
This paper proposes a rule-enhanced legal judgment prediction framework based on first-order logic (FOL) formalism and comparative learning (CL) to develop an adaptive adjustment mechanism for legal judgment logic and further enhance performance in LJP. 
Inspired by the process of human exam preparation, our method follows a three-stage approach: first, we initialize judgment rules using the FOL formalism to capture complex reasoning logic accurately; next, we propose a Confusion-aware Contrastive Learning (CACL) to dynamically optimize the judgment rules through a quiz consisting of confusable cases; finally, we utilize the optimized judgment rules to predict legal judgments. 
Experimental results on two public datasets show superior performance across all metrics. The code is publicly available\footnote{https://anonymous.4open.science/r/RLJP-FDF1}.
\end{abstract}

\section{Introduction}

The application of AI in law is growing annually, demonstrating capabilities in both assisting judicial professionals and providing legal consultation services to the public.
Legal Judgment Prediction (LJP) aims to predict the outcome of a case based on its facts, typically including the applicable legal provisions, accusation, and prison terms. As legal judgments are highly professionalized, legal knowledge is essential for LJP predictions. Researchers often employ deep learning models enhanced with legal knowledge or judicial precedents to achieve LJP.

The current methods of LJP mainly consist of two approaches: modeling legal knowledge and retrieving judicial precedents or legal knowledge. (1) Methods modeling legal knowledge typically use neural networks to model legal provisions and extract semantic associations between case facts and legal knowledge\cite{XuWCPWZ20, RN31}.
While these models can efficiently extract relevant legal knowledge for LJP relying on semantic similarity, they overlook the intrinsic logic of cases. 
(2) Other methods leveraged semantic techniques to search for related judicial precedents or legal knowledge\cite{RN33, RN35}. 
Judicial precedents and legal knowledge as references to enhance the model's understanding of the case circumstances, but they tend to study similar cases or knowledge instead of modeling the logical reasoning process. 
In summary, the above two types of methods rely on textual and semantic matching but ignore capturing the logic of judging legal cases, where logic in reasoning is crucial in legal judgment. 

To address the aforementioned issues, researchers applied legal judgment logic to enhance reasoning in traditional deep learning models or large language models (LLMs) for LJP. 
Some researchers employ deep learning models to extract logical premises from complex cases \cite{Yue0JWZACYW21, RN38} or handle the legal judgment using rules defined by experts \cite{gan2021judgment}. 
The integration of legal judgment logic has improved the reasoning capabilities of deep learning models. However, the ability of deep learning models is still limited by their model capacity and training data scale. 
As LLMs have remarkable abilities in text comprehension and logical reasoning, researchers combined judgment logic with LLMs for LJP.
Researchers\cite{deng2023syllogistic} use LLMs to retrieve relevant legal provisions and extract four types of criminal elements from case facts for judgment based on the classical legal logic syllogism. 
Prompts constructed by keyword structure from legal experts' determination guide LLMs to classify legal cases\cite{izzidien2024llm}. 
In summary, while existing methods using legal logic effectively analyze cases through established rules, their logic rigidity limits adaptation to case-specific contexts in complex scenarios. 

In this paper, we propose a dynamically optimized judgment rule method with adaptive adjustment mechanisms, formalizing rules(\S\ref{sec:3.2}) through First-Order Logic (FOL) - a symbolic language for complex legal reasoning. Our framework integrates tree-splitting operations with contrastive learning, where confusable cases systematically generate correct/erroneous judgment pairs for Confusion-Aware Contrastive Learning (CACL) (\S\ref{sec:3.3}). Iterative optimization produces enhanced rules with superior discriminative power in complex cases, enabling LLMs to achieve precise judgment predictions. This culminates in \textbf{RLJP} (\textbf{R}ule-enhanced \textbf{L}egal \textbf{J}udgment \textbf{P}rediction), a logic-enhanced system operationalizing FOL rules for legal judgment prediction (\S\ref{sec:3.4}).

Our contributions are: 
(1) We propose a dynamic rule optimization method. Pioneer the modeling of judgment rules optimization as tree splitting, and use the CACL mechanism for self-adaptive rules, overcoming the limitations of fixed rules in handling complex cases; 
(2) We propose RLJP, a method that novelly integrates FOL judgment rules to enhance reasoning ability for LJP; 
(3) Comprehensive evaluations on two public datasets demonstrate state-of-the-art performance across all metrics compared to baseline methods. To facilitate future research, we make our code publicly available.


\section{Related Work}

LJP aims to predict judicial outcomes by employing traditional models or LLMs to analyze legal cases \cite{10216994, RN30, zhang2024hd}. Broadly, LJP can be divided into two main approaches: one based on semantic similarity \cite{liu2024semdr} and the other guided by judgment logic \cite{cheng2024research}.
\subsection{LJP Assisted by Semantic Similarity}
The semantic similarity-assisted method for LJP involves modeling domain knowledge and applying retrieval techniques \cite{zhang2023case}. 

The former mainly utilizes neural networks to model legal knowledge \cite{Yue0JWZACYW21, yao2024lawyer} for feature extraction and to establish semantic associations among the knowledge.
To improve the accuracy of crime prediction, the prevailing methods primarily focus on refining the attention mechanism \cite{10216994, yu2023enhancing, he2025simulating}. Additionally, some methods infuse domain-specific legal knowledge into LLMs \cite{wang2024legalreasoner, RN44}, thereby enhancing the models' capability to identify and leverage critical information. Regarding confusion issue in criminal charges,  \citeauthor{zhang2023contrastive} (\citeyear{zhang2023contrastive}) and \citeauthor{gan-etal-2023-exploiting} (\citeyear{gan-etal-2023-exploiting}) utilized contrastive learning to capture fine-grained distinctions.
Using retrieval techniques refers to utilizing search engines \cite{RN33, RN34, RN41} to obtain external knowledge from semantically similar judicial precedents. \citeauthor{RN35} (\citeyear{RN35}) introduced the retrieval technology into the SimCourt multi-agent framework. \citeauthor{RN47} (\citeyear{RN47}) constructed a hierarchical semantic ID for the retrieved candidate documents.

Legal judgment fundamentally relies on rigorous, logic-bound reasoning demanding systematic analysis. This process necessitates precise legal interpretation, detailed case fact scrutiny, and accurate application of legal principles, whereas solely relying on textual semantic similarity fails to ensure the accuracy and logical rigor required in LJP.

\subsection{LJP Assisted by Case Judgment Logic}
The legal judgment logic-enhanced method for LJP leverages models to summarize the knowledge embedded in case details \cite{chen2020legal, yang2022mve} and extract the underlying reasoning logic.

Existing methods focus on using the entire factual description \cite{ chai2020description, hong2023improving} to generate judgment results, overlooking the actual judicial scenario where judges consider various criminal circumstances \cite{wu2022towards} to determine the verdict and sentencing. Consequently, \citeauthor{Yue0JWZACYW21} (\citeyear{Yue0JWZACYW21}) proposed a legal judgment framework, NeurJudge, which can divide facts and the prediction results of intermediate subtasks into different scenarios. \citeauthor{RN38} (\citeyear{RN38}) further proposed NeurJudge+, which integrates the semantic labels of accusations and legal provisions into facts. \citeauthor{gan2021judgment} (\citeyear{gan2021judgment}) injected legal knowledge as an additional logical rule into the neural network.  \citeauthor{deng2023syllogistic} use LLMs to retrieve relevant legal provisions and extract four types of criminal elements from case facts for judgment based on the classical legal syllogism. Prompts constructed by keyword structure from legal experts' determination guide LLMs to classify legal cases\cite{izzidien2024llm}.


In summary, the inherent structural rigidity of previous methods limits their ability to dynamically adapt to diverse, case-specific logical frameworks. Thus, when confronted with complex legal cases, these inflexible frameworks struggle to reconcile conflicting evidence or interpretative contradictions.  In light of this, this paper aims to develop an adaptive adjustment mechanism for legal judgment logic, thereby further enhancing LJP performance in complex and contradictory situations.

\section{Method}

\subsection{Task Definition}
In this paper, we focus on the problem of legal judgment prediction. First, we clarify the definition of the terms as follows:
\begin{itemize}
\item \textbf{Judgment} is the final decision made by a judge in a legal case based on the facts and the judgment rule. It typically consists of the law article, the charge, and the prison term. We represent the judgment of a case as $j = (article, charge, prison\_term)$, where $(article, charge, prison\_term)$ refers to the labels of provisions, accusation, and prison term, respectively.

\item \textbf{Precedent} is the previous case with a similar fact. For a given judgment label, there can be several precedents, which can be denoted as $S = \{s_1, s_2, ..., s_n\}$, where $n$ is the number of precedents.

\item \textbf{Rule} refers to formalized patterns derived from summarizing recurrent case development trajectories across similar judicial decisions in our study.
\end{itemize}

\textbf{Problem 1} Induction Rule.  Extract causal factors from case facts for the content of the rule, and construct the confusable case fact to optimize the rule.

\textbf{Problem 2} Legal Judgment Prediction. Given the case fact $f$, our task is to analyze $f$ based on $rule$, then predict judgment $j = (article, charge, prison\_term)$.

\subsection{Model Overview} \label{sec:3.1}

\begin{figure*}[]
  \centering
  \includegraphics[width=16cm]{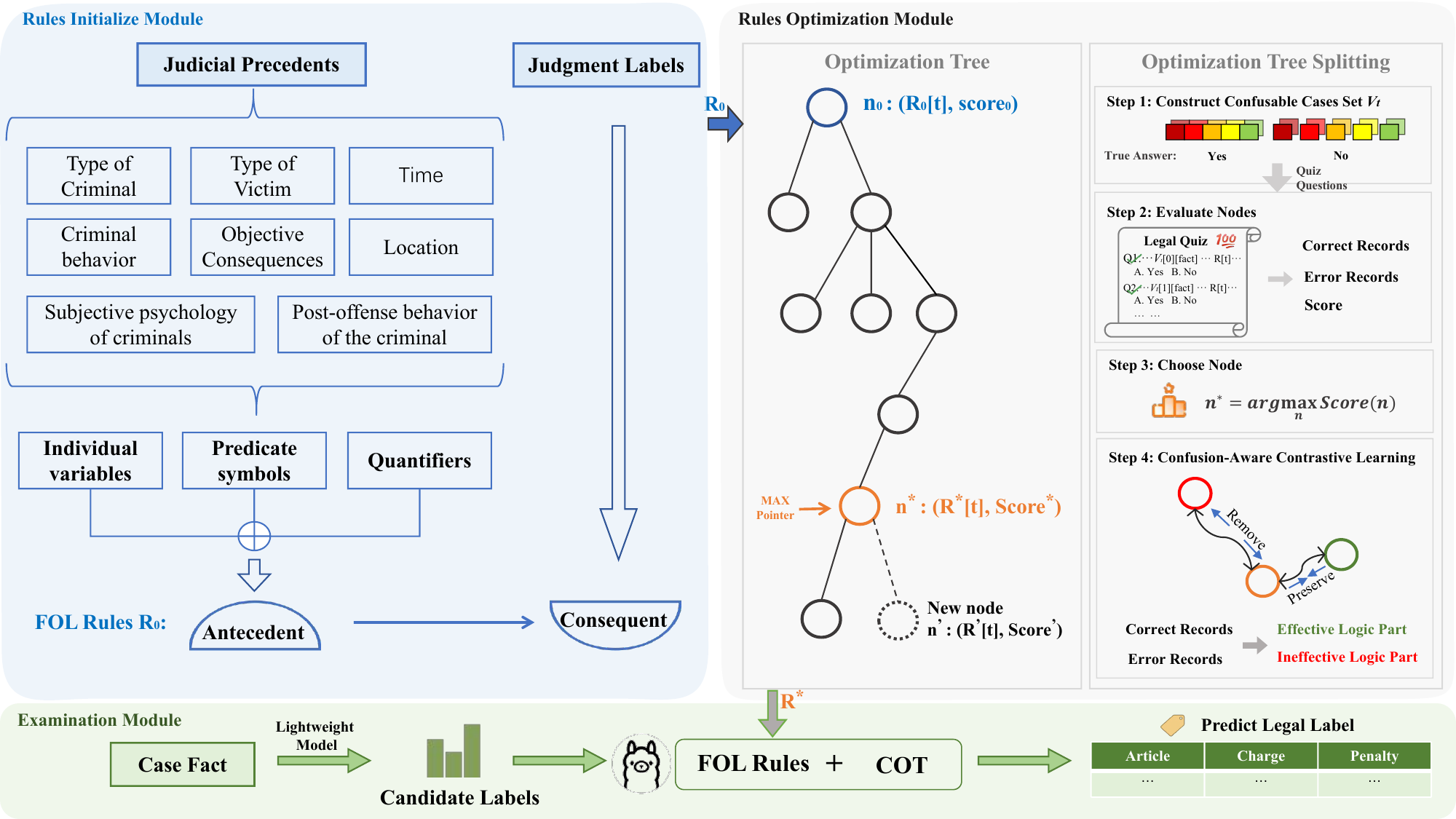} 
  \caption{RLJP framework. 
  The blue box is the Rule Initialization Module on the upper left(\S\ref{sec:3.2}), the gray box is the Rule Optimization Module(\S\ref{sec:3.3}) with Confusion-Aware Contrastive Learning, and the green box is the Examination Module(\S\ref{sec:3.4}) for the completion of LJP tasks based on the optimized rules.
  }
  \label{fig: method}
\end{figure*}

During the learning process of human students, they typically first acquire examples and knowledge from textbooks, then summarize problem-solving logic or methods. These methods are later tested through quizzes to assess their accuracy, and the final learning outcomes are evaluated in the final exam.  
Inspired by this, we propose the RLJP framework (Figure \ref{fig: method}) to enhance reasoning ability for the LJP task.
The framework contains three modules. 
\begin{itemize}
    \item \textbf{Rules Initialization Module} (\S \ref{sec:3.2}). This module aims to utilize an LLM agent to generate rude judgment rules in the FOL formalism, leveraging legal provisions and precedents.  This process mirrors how students acquire knowledge and summarize problem-solving logic.
    \item  \textbf{Rules Optimization Module}  (\S \ref{sec:3.3}). This module aims to use the Confusable Case Set to optimize the FOL judgment rules through Confusion-Aware Contrastive Learning (CACL), which is similar to how students optimize their problem-solving logic through quiz experience. 
    \item \textbf{Examination Module} (\S \ref{sec:3.4}). This module uses an LLM agent to predict legal judgments by applying optimized judgment rules in combination with candidate labels generated by a lightweight model, which mirrors the final exam of the students. 
\end{itemize}

\subsection{Rules Initialization} \label{sec:3.2}

Inspired by the cognitive process where students first encounter a specific problem type and subsequently extract solution logic, we develop reasoning rules for determining categories of legal judgments through context learning. To precisely describe the reasoning rules, we utilize the FOL formalism, which can accurately express complex logic. 

Each FOL judgment rule $Rule: A\to C$ comprises an antecedent $A$ and a consequent $C$. The content of $A$ is determined by the circumstance logic. And $C$ is constructed by one or two legal judgment labels, including $article$, ($article$, $charge$), and ($article$, $prison\_term$).

Since legal provisions offer a clear legal basis for accusation determination, we select precedents violating the same legal provisions and receiving the same accusation to generate accusation judgment rules. As different legal provisions set different sentencing standards, we extract precedents violating the same legal provisions and receiving the same prison term to help generate prison-term judgment rules. This combination reminds the LLM agent that the content of judgment rules should strictly adhere to established legal provisions when initializing rules.

To systematically summarize the underlying logical patterns in case developments of similar legal cases, we employ FOL symbols to formalize legal judgment rules through a three-step process.

\textbf{(1) Summarize circumstance logic.} This step aims to analyze some causal factors from similar legal cases. Specifically, we use the LLM agent to summarize causal factors causing the legal judgment from some precedents in the same legal judgment, including the category of criminal subject, the category of the victim, the time and location of the crime, the criminal behavior, the objective consequences of the crime, and the criminal's subjective mental state. These factors are crucial for constructing the antecedent $A$ of the FOL judgment rules. 

\textbf{(2) Define FOL symbols.} The objective of this step is to establish the logical relationship between the causal factors and the legal judgment in the case. 
For this purpose, we use the LLM agent to define causal factors with FOL symbols, including variables, predicate symbols, and quantifiers. 
Variables can represent the entities, time, location, events, consequences, and mental state in factors. 
Predicate symbols describe specific attributes of the criminal subject, the victim, and the criminal behavior. 
The universal quantifier $\forall$ and the existential quantifier $\exists$, express the rule's generality or specificity. 

\textbf{(3) Construct FOL judgment rules.} The purpose of this step is to standardize the format of rules to facilitate subsequent optimization and reasoning processes. The judgment rules are constructed using the antecedent $A$ and the consequent $C$. $Rule: A\to C$ in this step. The antecedent $A$ comprises multiple FOL symbols and logical connectives ($\vee$, $\wedge$, and $\neg$). The consequent $C$ is the same as LJP labels, including legal provisions $article$, (legal provisions $article$, accusation $charge$), and (legal provisions $article$, prison terms $prison\_term$).

\subsection{Confusion-Aware Rule Optimize Engine} \label{sec:3.3}

In RLJP, we propose a Confusion-Aware Rule Optimization Module to eliminate ambiguous boundaries among rules,  which can learn from correct and incorrect experiences in confusable cases reasoning by iteratively optimizing rules with CACL. The full procedure of this Optimization is described in Algorithm \ref{arg:module}. 
Specifically, this module works in two-stage steps: construction confusable case set(\S \ref{sec:3.3.1}), and rules optimization(\S \ref{sec:3.3.2}) depends on Confusion-Aware Contrastive Learning(CACL).
\begin{algorithm}
\footnotesize
\caption{Confusion-Aware Rule Optimization}
\label{arg:module}
\begin{algorithmic}[1]
    \REQUIRE $Precedents$: Precedent set;
    \REQUIRE $target$: Target label which is the consequence of the rules;
    \REQUIRE $RULE$: Initial rule set ;
    \REQUIRE $DefinedScore$: A predefined score threshold serves as the stopping criterion for terminating the iterative updating process;
    \REQUIRE $Num$: The required number of confusable cases.
    \STATE $S_{postive} \leftarrow Precedents[target]$
    \STATE $S_{other} \leftarrow Precedents[other]$
    \STATE $S_{negative} \leftarrow RankSimilarity(S_{postive},S_{other}, Num)$
    \STATE Confusable case set $V_{target} \leftarrow S_{negative} \cup S_{negative}$
    \STATE $r_{0} \leftarrow RULE[target] $
    \STATE $MaxScore \leftarrow 0, MaxPointer \leftarrow Null $
    \STATE $R_{tree} \leftarrow [r_{0}] $
    
    \FOR{r in $R_{tree}$}
        \IF{Is\_Evaluate($r$)} 
                \STATE ($score$, reasoning records) $\leftarrow  history(r)$
            \ELSE
                \STATE ($score$, reasoning records) $ \leftarrow Evaluate(V_{target}, r)$
            \ENDIF
            \IF{$score > MaxScore$} 
                \STATE $ MaxScore \leftarrow  score$
                \STATE $ MaxPointer \leftarrow  r$
                \STATE $ R ^* \leftarrow MaxPointer$
            \ENDIF 
        \IF{$MaxScore >= DefinedScore$}
            \STATE break;
        \ENDIF
        \STATE $R' \leftarrow CACL(R^*,$ reasoning records $)$
        \STATE $R_{tree} \leftarrow$ Add\_ChildNode($R',R^*$)
    \ENDFOR
    
    \RETURN  $R^{*}$;
\end{algorithmic}
\end{algorithm}

\subsubsection{Construction of the Confusable Cases Validation Set} \label{sec:3.3.1}
\label{sec:vtarget}

The confusable case set is defined as cases characterized by highly similar factual circumstances but differing legal outcomes in our paper.
In the real world, confusable cases often pose significant challenges to professionals tasked with making judgment decisions. Thus, we construct the confusable case set as a benchmark to evaluate the quality of legal rules generated in RLJP. 
Specifically, to measure the similarity between cases, we use the Bi-directional Generative Encoder(BGE) model to generate vector representations of case fact texts and calculate the cosine distance between vectors. 
The construction process of this set is as follows. 

To find confusable cases for legal judgment $target$, we first find the case set, whose legal judgment label is $target$, denoted as $S_{positive}$, and the case set for other legal judgment labels, denoted as $S_{other}$. Then, to asses the similarity score between $S_{positive}$ and $S_{other}$, we convert each case fact set into fixed-dimensional embedding vectors using the BGE model: $E_{t}=\left \{ E_{t}^{1}, E_{t}^{2},..., E_{t}^{m}  \right \} $ for $S_{positive}$ and $E_{o}=\left \{ E_{o}^{1}, E_{o}^{2},..., E_{o}^{n}  \right \} $ for $S_{other}$, $E_{t}^{i}, E_{o}^{i}\in R ^{d}$. 
Then, we compute the cosine similarity between $E_{t}$ and $E_{o}$. The results are organized into a similarity matrix $A$, where $A(i,j)=\frac{E_{t}^{i}\cdot  E_{o}^{i}}{\left \| E_{t}^{i} \right \|  \left \|  E_{o}^{i} \right \| }$ represents the similarity score of two cases $S_{positive}^i$ and $S_{other}^j$. 

The higher the similarity score between cases in different judgment labels means that they contain similar crime circumstances, the more likely they are to be confusable for legal judgment. 
For this reason, we sort the similarity matrix $A$ in descending order based on the similarity scores of each row, and the first column is extracted to obtain a set of negative sample cases for judgment $target$: $S_{negative} \in S_{other}$. Finally, the confusable case set $V_{target}$ can be constructed: $V_{target}=S_{negative} \cup S_{positive}$.


\subsubsection{Optimization Tree Splitting} 
\label{sec:3.3.2}
To iteratively optimize legal judgment rules, we formulate the process of rule optimization as the process of tree-splitting. These nodes of the optimization tree are different versions rule during optimization. 
In one iteration, we first create some reasoning quizzes constructed by the confusable cases set, and calculate the score of the judgment rules for best-first splitting. And then, we collect correct and incorrect reasoning experience in quizzes for CACL, which autonomously optimizes rules to keep effective logic parts and update ineffective logic parts by analyzing experience.



\textbf{Optimization Tree Definition.} We formalize judgment rule optimization as a weighted tree structure $ T_{target} = (N, E, W) $, where $ N $ denotes rule in different versions, $ E \in N*N $ encodes rule optimization relationships, and $W: N \to [0,1]$ quantifies rules' performance using quiz score using the confusable case set. The root node $n_{0} \in N$ corresponds to the initial rule formulation from \S \ref{sec:3.2}, with directed edges $(n_p,n_c) \in E$ representing optimization paths where child rule $n_c$ evolves from parent $n_p$ through experience-based refinements. Node weights $W(n)$ establish a partial ordering over rule versions, enabling performance-guided choice in the optimization cycle.

\textbf{Optimization Tree Splitting.} Legal judgment rules optimization contains three iterative steps. 

The first step aims to evaluate and choose the node. To select the basis node for optimization, in this step, we evaluate all rules in the current tree using the confusable case set, and choose the highest accuracy node $n^{*}$ (rule $R^{*}$) to mark with the MAX pointer. We propose the CACL method to optimize $R^{*}$. The CACL method finds optimization directions based on evaluation experience and generates optimized rules. 
Finally, the third step aims to integrate the node. Attach the new rule as a child node to $n^{*}$, repeating these three phases until either a predefined accuracy threshold is achieved by the optimized rule or the maximum iteration count is reached. 
Detailed explanations of each step are as follows.

\textbf{Step 1: Evaluate and Choose Node.} Inspired by staged assessments in educational settings, we design a quiz using confusable cases to evaluate the rules. 
First, we format case facts from the $V_{target}$ into single-choice quiz questions. The template for the construction of single-choice questions is provided in Figure \ref{fig: ljp}. Then the LLM agent using $R_n$ chooses a predicted option and generates a reasoning process for each question. The capability of $R_n$ is assessed based on the accuracy of predicted options. 
The experience of choosing correctly consists of True Positives (TP) and True Negatives (TN). 
The experience of choosing incorrectly consists of False Positives (FP) and False Negatives (FN). The weight of node $W(n)$ is computed as formalized in Equation \ref{eq:wegit_caculation}.

\begin{equation}
  \label{eq:wegit_caculation}
  W(n)=\frac{TP+TN}{TP+TN+FP+FN} 
\end{equation}

The selected node $n^{*} = arg\max_{n\in N}W(n)$ is the node with the highest weight in the tree, corresponding to the rule $R^{*}$ that achieves the highest accuracy on the validation set.

\begin{algorithm}
\footnotesize
\caption{Confusion-Aware Contrastive Learning}
\label{arg: CACL}
\begin{algorithmic}[1]
    \REQUIRE $R^{*}$: Current rule (Anchor);
    \REQUIRE $Correct(R^{*})$: Set of positive samples (valid reasoning records);
    \REQUIRE $Incorrect(R^{*})$: Set of negative samples (invalid reasoning records).
    \STATE $\bigtriangledown{keep}  \leftarrow LLM(R^*,Correct(R^{*}))$
    \STATE $\bigtriangledown{imp}  \leftarrow LLM(R^*,Incorrect(R^{*}))$    
    \STATE $\bigtriangledown{R}  \leftarrow LLM(\bigtriangledown{keep},\bigtriangledown{imp})$
    \STATE $R'  \leftarrow LLM(\bigtriangledown{R},R^*)$
    
    \RETURN $R'$
\end{algorithmic}
\end{algorithm}

\textbf{Step 2: CACL Method.} To optimize the current optimal rule $R^*$, we propose the CACL to generate an optimized new rule, which simulates students reflecting on the correct and incorrect problem-solving process. The full procedure of the CACL method is described in Algorithm \ref{arg: CACL}, which consists of three core steps as follows. 

\textbf{Step 2-1: Construct triplets.} CACL constructs the experience of rule evaluation as contrast triplets. The key triplet in CACL is (Anchor $R^{*}$, Positive Samples $Correct(R^*)$, Negative Samples $Incorrect(R^*)$). Anchor is the current rule $R^*$. Positive samples are correct reasoning records $Correct(R^*)$ (TP and TN) in evaluation, which contain quiz questions, reasoning process, correct option, and predicted Option. Negative samples are composed of the same quadruples and use wrong reasoning records $Incorrect(R^*)$ (FP and FN) in the evaluation.
The quadruple construction is provided in Figure \ref{fig: quiz_experience}.

\textbf{Step 2-2: Generate optimization direction.} CACL analyzes contrast triplets for optimization, the LLM agent generates optimization direction $\bigtriangledown R$ of $R^{*}$ based on the positive and negative triplets, including effective logic parts $\bigtriangledown{keep}$ and ineffective logic parts $\bigtriangledown{imp}$. $\bigtriangledown R =(\bigtriangledown{keep},\bigtriangledown {imp})$. The prompt template for generating optimization direction is shown in Figure \ref{fig: optDirection}.

\textbf{Step 2-3: Optimize rules.} The optimization direction generated from CACL guides the LLM agent to maintain the effective logic part and improve the ineffective logic part. The prompt template for the rule optimization is detailed in Figure \ref{fig: optRule}.

\textbf{Step 3: Integrate Node.} The optimized $R'$ is added to the tree as a child node of $R^{*}$. We continue to optimize rules until the optimized rule achieves the predefined accuracy or the maximum iteration count is reached.

\subsection{Examination} \label{sec:3.4}
Inspired by the final examination process after many quizzes in educational settings, this module executes the LJP task with the optimized rules. In this module, we start using a lightweight BERT model to output the top 10 probable legal labels. Then we apply FOL judgment rules against each candidate label with the Chain-of-Thought method to predict legal judgment. 
The prompt template for completing the LJP task, as Figure \ref{fig: quiz_experience} shows which is the same as the quiz template. If none of the candidates meet the logical constraints, we initiate a stochastic traversal of the remaining labels based on their rules. Furthermore, we activate the abstract template as Figure \ref{fig: abstract} shows to generate the abstract of the case fact when the length of the fact exceeds a defined threshold. The abstract template aims to generate condensed case abstracts that retain legally relevant features while eliminating redundant details.


\section{Experiments}
\subsection{Settings}
\textbf{Datasets.} Following previous LJP works \citep{ZhongGTX0S18, XuWCPWZ20, Yue0JWZACYW21, WuZL0LZS0K23}, we conduct experiments on datasets CAIL2018 \footnote{\url{https://github.com/china-ai-law-challenge}} and CJO22\citep{WuZL0LZS0K23}. For the CAIL2018 dataset, we randomly divide it into training, validation, and test sets with a ratio of 8:1:1\citet{WuZL0LZS0K23}. For the CJO22 dataset, following \citet{WuZL0LZS0K23}, we use it solely as an additional test set. 
The table in Appendix \ref{sec:appendix_dataset} presents the details of two datasets.

\noindent\textbf{Evaluation Metrics.} We use Accuracy (Acc), Macro-Precision (Ma-P), Macro-Recall (Ma-R), and Macro-F1 (Ma-F) as the evaluation metrics following previous LJP works \citep{ZhongGTX0S18, XuWCPWZ20, Yue0JWZACYW21, WuZL0LZS0K23}.

\noindent\textbf{Baselines.} We compare with: \textbf{(1) CNN}\citep{LeCunBDHHHJ89} use different kernel convolutional operations to extract text features for classification; 
\textbf{(2) BERT}\citep{DevlinCLT19} can be easily fine-tuned on downstream tasks such as LJP; 
\textbf{(3) TopJudge}\citep{ZhongGTX0S18} employs multi-task learning and captures the dependencies among the three sub-tasks in LJP; 
\textbf{(4) R-Former}\cite{DongN21} formalizes LJP as a node classification problem over a global consistency graph; 
\textbf{(5) LADAN\cite{XuWCPWZ20}} uses graph distillation to extract discriminative features of the fact;
\textbf{(6) NeurJudge}\citep{Yue0JWZACYW21} splits the fact description into different parts for making predictions; 
\textbf{(7) EPM}\cite{DBLP:conf/ijcai/FengL022} locates event-related information essential
for judgment;
\textbf{(8) CTM}\citep{LiuDLP022} uses case triple modeling from contrastive case relations; 
\textbf{(9) PLJP}\citep{WuZL0LZS0K23} uses the collaboration between LLMs and domain-specific models for LJP; 
\textbf{(10) Llama3}\citep{WuZL0LZS0K23}, which is Meta's advanced open-source language model; 
\textbf{(11) D-LADAN}\citep{RN31} solves LJP confusion via graph and memory mechanisms.

\noindent\textbf{Implementation Details.} We directly adopted Bert-base-chinese\footnote{\url{https://huggingface.co/google-bert/bert-base-chinese}} for candidate labels, Gpt-4o for rules optimization, and Llama3-chinese\cite{Llama3-Chinese} for other steps because the language of the cases in the data set is Chinese. For the length limit, we use three precedents in Section \ref{sec:3.2}. For the parameter settings of the baseline, we strictly follow the original paper. More details are in Appendix \ref{sec:appendix_set}.

\begin{table*}[]
    \centering
    \setlength{\tabcolsep}{0.65mm}{
    \fontsize{9.0}{12}\selectfont
    \caption{Experiment results of LJP in CAIL2018 dataset. “\textbf{Bold}” indicates optimal results, and “\underline{underline}” indicates sub-optimal results. The experimental results represent the average values obtained from five test rounds. }
\label{tab:cail}
\footnotesize
\begin{tabular}{cc|cccccc}
\hline
\multicolumn{2}{c|}{\multirow{2}{*}{\textbf{Method}}}   & \multicolumn{2}{c}{\textbf{Law Article}} & \multicolumn{2}{c}{\textbf{Charge}} & \multicolumn{2}{c}{\textbf{Term-of-Penalty}} \\ \cline{3-8} 
\multicolumn{2}{c|}{}                                   & \textbf{Acc}        & \textbf{Ma-F}      & \textbf{Acc}     & \textbf{Ma-F}    & \textbf{Acc}        & \textbf{Ma-F}      \\ \hline
\multirow{2}{*}{Navie Model}    & BERT            & 82.77               & 35.82              & 89.1             & 89.63            & 40.00               & 33.58              \\
                                & Llama3-Chinese & 2.62                & 1.11               & 22.63            & 10.04            & 12.00               & 10.53              \\ \hline
\multirow{4}{*}{Semantic-based}  & Neurjudge      & 76.91               & 53.56              & 82.13            & 82.36            & 33.53               & 36.53              \\
                                & PLJP\_Bert    & 87.07               & \underline{56.63}              & {\ul 94.99}      & {\ul91.33  }          & {\ul 48.72}         & 35.43              \\
                                & D-LADAN        & \textbf{91.05}         & \textbf{ 58.38}        & 90.82            & 62.08            & 38.31               & 25.00              \\
                                & KnowJudge       & --                  & --                 & 89.00            & 85.00            & --                  & --                 \\ \hline
\multirow{2}{*}{Logic-based}    & EPM           & 85.80               & 47.32              & 91.20            & 90.46            & 40.25               & 37.34              \\
                                & CTM           & 84.72               & 45.10              & 90.28            & 86.30            & 39.56               & {\ul 37.84}        \\ \hline
\multicolumn{2}{c|}{\textbf{Ours(RLJP)}}                               & \underline{88.12}      & 54.42     & \textbf{96.00}   & \textbf{96.10}   & \textbf{54.72}      & \textbf{48.45}     \\ \hline
\end{tabular}}
\end{table*}
\nopagebreak

\begin{table*}[]
    \centering
    \setlength{\tabcolsep}{0.65mm}{
    \fontsize{9.0}{12}\selectfont
    \caption{Experiment results of LJP in CJO22 dataset. \textbf{Bold} and \underline{underlined} denote the same notation as Tab. \ref{tab:cail}. }
\label{tab:cjo22}
\footnotesize
\begin{tabular}{cc|cccccc}
\hline
\multicolumn{2}{c|}{}                                  & \multicolumn{2}{c}{\textbf{Law Article}}        & \multicolumn{2}{c}{\textbf{Charge}}             & \multicolumn{2}{c}{\textbf{Term-of-Penalty}}        \\ \cline{3-8} 
\multicolumn{2}{c|}{\multirow{-2}{*}{\textbf{Method}}} & \textbf{Acc}                       & \textbf{Ma-F}                      & \textbf{Acc}                       & \textbf{Ma-F}                      & \textbf{Acc}                       & \textbf{Ma-F}                      \\ \hline & BERT    & 82.62  & 45.83 & 80.5  & 78.36  & 36.8  & 27.03   \\
\multirow{-2}{*}{Navie model}   & Llama3-Chinese   & 3.01   & 1.36  & 27.72  & 28.28 & 15.66    & 15.56                              \\ \hline
& Neurjudge & 71.38  & 52.62     & 71.85   & 68.66   & 26.8   & 25.97   \\
& PLJP\_Bert   & {\ul 94.18}   & {\ul 74.84}   & {\ul 94.18}  & {\ul89.05}  & 43.52     & {\ul 31.98}    \\
\multirow{-3}{*}{Semantic-based}            & D-LADAN        & 90.65   & 44.49  & 88.95     & 28.48   & {\ul 46.18}   & 22.73      \\ \hline
                                            & EPM                       & 84.19                              & 44.39                              & 83.49                              & 81.87                              & 36.91                              & 30.2                               \\
\multirow{-2}{*}{Logic-based}               & CTM                      & 79.44                              & 43.43                              & 79.33                              & 82.81                              & 36.81                              & 26.46                              \\ \hline
\multicolumn{2}{c|}{\textbf{Ours(RLJP)}}                                                      & \multicolumn{1}{l}{\textbf{94.55}} & \multicolumn{1}{l}{\textbf{91.28}} & \multicolumn{1}{l}{\textbf{96.12}} & \multicolumn{1}{l}{\textbf{96.83}} & \multicolumn{1}{l}{\textbf{48.50}} & \multicolumn{1}{l}{\textbf{49.18}} \\ \hline
\end{tabular}
\label{tab:cjo22}}
\end{table*}

\subsection{Main Results} \label{sec: main_exp}

Tables \ref{tab:cail} and Table \ref{tab:cjo22} show the performance comparison results of RLJP and various baseline models on the CAIL2018 and CJO22 datasets, respectively, which are the average values obtained from five test turns.
The experimental results show that RLJP achieved optimal performance in all metrics, verifying the enhancing effect of FOL judgment rules on LJP. Specifically, compared to the suboptimal model, RLJP achieved an average improvement of 1.43\% in Acc and 14.98\% in Ma-F on the CAIL2018 and CJO22 datasets. These experimental results fully validate the effectiveness and superiority of our method in LJP, indicating its significant advantages in legal case judgment. 

It is worth noting that compared to the tasks of predicting legal provisions and criminal charges, the task of prison term prediction still presents significant challenges, mainly reflected in its relatively small performance improvement and lower overall prediction accuracy compared to the other two tasks. This phenomenon may be related to the more complex sentencing factors and subjective judgments of judges involved in sentence prediction and is worth further exploration and improvement in subsequent research.

\setlength{\tabcolsep}{1mm}
\begin{table}[]
\centering
\renewcommand{\arraystretch}{1.2}
\caption{Results of ablation experiments on CAIL2018. “w/o R”, “w/o Optimize”, “w/o CACL”, and “w/o Candidate” denote removing judgment rules, optimization modules, CACL method,
and candidate labels, respectively.}
\fontsize{8.5}{12}\selectfont
\label{tab: aba_cail}
\begin{tabular}{l|cccccc}
\hline
\multicolumn{1}{c|}{\multirow{2}{*}{\textbf{Method}}}  &  \multicolumn{2}{c}{\textbf{Law Article}} & \multicolumn{2}{c}{\textbf{Charge}}  & \multicolumn{2}{c}{\textbf{Prison Term}}  \\ \cline{2-7} 
         \multicolumn{1}{c|}{} &   \multicolumn{1}{c}{Acc}   & \multicolumn{1}{c}{Ma-F} & \multicolumn{1}{c}{Acc} & \multicolumn{1}{c}{Ma-F} & \multicolumn{1}{c}{Acc} & \multicolumn{1}{c}{Ma-F} \\
\hline
w/o R & 18.92 & 26.53 & 82.43 & 81.25 & 28.05 & 16.89  \\
w/o Optimize & 85.9 & 82.98 & 83.13 & 84.28 & 39.74 & 41.16 \\
w/o CACL & 86.58 & 82.31 & 87.91 & 86.2 & 20.81 & 24.28  \\
w/o Candidate & 37.25 & 30.17 & 45.64 & 35.74 & 17.45 & 19.00 \\
\hline
\textbf{ours(RLJP)} & \textbf{91.27} & \textbf{88.32} & \textbf{96.00} & \textbf{96.10} & \textbf{54.72} & \textbf{48.45}\\
\hline
\end{tabular}
\end{table}

\setlength{\tabcolsep}{1mm}
\begin{table}[]
\centering
\renewcommand{\arraystretch}{1.2}
\caption{Results of ablation experiments on CJO22. “w/o R”, “w/o Optimize”, “w/o CACL”, and “w/o Candidate” denote removing judgment rules, optimization modules, CACL method,
and candidate labels, respectively.}
\fontsize{8.5}{12}\selectfont
\label{tab: aba_cjo}
\begin{tabular}{l|cccccc}
\hline
\multicolumn{1}{c|}{\multirow{2}{*}{\textbf{Method}}}  &  \multicolumn{2}{c}{\textbf{Law Article}} & \multicolumn{2}{c}{\textbf{Charge}}  & \multicolumn{2}{c}{\textbf{Prison Term}}  \\  \cline{2-7} 
         \multicolumn{1}{c|}{} &   \multicolumn{1}{c}{Acc}   & \multicolumn{1}{c}{Ma-F} & \multicolumn{1}{c}{Acc} & \multicolumn{1}{c}{Ma-F} & \multicolumn{1}{c}{Acc} & \multicolumn{1}{c}{Ma-F} \\
\hline
w/o R  & 14.31 & 15.59 & 71.08 & 68.71 & 20.12 & 18.17 \\
w/o Optimize & 85.44 & 83.98 & 89.7 & 89.00 & 37.35 & 37.52 \\
w/o CACL & 84.98 & 84.08 & 91.8 & 90.36 & 31.52 & 32.63 \\
w/o Candidate & 21.05 & 15.06 & 71.58 & 68.86 & 23.53 & 23.17 \\ 
\hline
\textbf{ours(RLJP)} & \textbf{94.55} & \textbf{91.28} & \textbf{96.40} & \textbf{96.58} & \textbf{48.50} & \textbf{49.18}\\
\hline
\end{tabular}
\end{table}

\subsection{Ablation Experiment}

To systematically evaluate the contributions of different components to the performance of LJP, we designed ablation experiments. Results of ablation experiments in Table \ref{tab: aba_cail} and Table \ref{tab: aba_cjo}, we can conclude that: 
(1) “w/o R” represents removing judgment rules (\S \ref{sec:3.2}) and causes the decrease of all metrics, which proves judgment rules greatly help the LLM agent in reasoning legal judgment. 
(2) We find removing the optimization module (“w/o Optimize”) causes the drop in metrics, proving that the dynamical Optimization module (\S \ref{sec:3.3}) is beneficial for improving the rules used in the process of judgment prediction. 
(3) In row 3 (“w/o CL”), results show some metrics drop and some metrics up when we optimize the rule based on the latest version rather than CACL (\S \ref{sec:3.3.2}). The up phenomenon shows the validity of optimization. The down phenomenon, caused by the content of the rule overfitting in some cases, shows that the performance-guided optimization method is important.
(4) “w/o Candidate” removes candidate labels in the Examination module (\S \ref{sec:3.4}), performing worse than RLJP. This indicates the importance of the lightweight model. 

\subsection{Analytical Experiment}

\setlength{\tabcolsep}{1mm}
\begin{table}[]
\centering
\renewcommand{\arraystretch}{1.2}
\caption{Results of analytical experiments on CAIL2018\_long. The experimental results represent the average values obtained from five test rounds.}
\fontsize{8.5}{12}\selectfont
\label{tab: ana_cail}
\begin{tabular}{l|cccccc}
\hline
\multicolumn{1}{c|}{\multirow{2}{*}{\textbf{Method}}}  &  \multicolumn{2}{c}{\textbf{Law Article}} & \multicolumn{2}{c}{\textbf{Charge}}  & \multicolumn{2}{c}{\textbf{Prison Term}}  \\ \cline{2-7} 
         \multicolumn{1}{c|}{} &   \multicolumn{1}{c}{Acc}   & \multicolumn{1}{c}{Ma-F} & \multicolumn{1}{c}{Acc} & \multicolumn{1}{c}{Ma-F} & \multicolumn{1}{c}{Acc} & \multicolumn{1}{c}{Ma-F} \\
\hline
PLJP(Bert) & 28.06 & 36.17 & 40.94 & 44.74 & 12.32 & 13.92 \\
\hline
\textbf{ours(RLJP)} & \textbf{79.27} & \textbf{78.27} & \textbf{91.67} & \textbf{91.81} & \textbf{35.14} & \textbf{36.23}\\
\hline
\end{tabular}
\end{table}

\setlength{\tabcolsep}{1mm}
\begin{table}[]
\centering
\renewcommand{\arraystretch}{1.2}
\caption{Results of analytical experiments on CJO22\_long. The experimental results represent the average values obtained from five test rounds.}
\fontsize{8.5}{12}\selectfont
\label{tab: ana_cjo}
\begin{tabular}{l|cccccc}
\hline
\multicolumn{1}{c|}{\multirow{2}{*}{\textbf{Method}}}  &  \multicolumn{2}{c}{\textbf{Law Article}} & \multicolumn{2}{c}{\textbf{Charge}}  & \multicolumn{2}{c}{\textbf{Prison Term}}  \\ \cline{2-7} 
         \multicolumn{1}{c|}{} &   \multicolumn{1}{c}{Acc}   & \multicolumn{1}{c}{Ma-F} & \multicolumn{1}{c}{Acc} & \multicolumn{1}{c}{Ma-F} & \multicolumn{1}{c}{Acc} & \multicolumn{1}{c}{Ma-F} \\
\hline
PLJP(Bert) & 38.93 & 58.87 & 42.64 & 44.71 & 23.26 & 22.29 \\
\hline
\textbf{ours(RLJP)} & \textbf{41.67} & \textbf{47.62} & \textbf{97.70} & \textbf{91.95} & \textbf{31.70} & \textbf{35.96}\\
\hline
\end{tabular}
\end{table}

To validate the performance advantages of RLJP in handling complex case facts, we selected the top 5\% of cases as the test cases based on case length, which contain more details and complex circumstances. The table in Appendix \ref{sec:appendix_long} presents the statistics of these two subsets. We compared the performance of the second-ranked PLJP (\S\ref{sec: main_exp}) with our RLJP. 

The experimental results in Table \ref{tab: ana_cail} and Table \ref{tab: ana_cjo} demonstrate that the proposed RLJP method is over-performing in judging complex facts. We can conclude that FOL judgment rules can effectively capture key elements in fact of cases, reducing interference from redundant information and focusing on critical facts that are decisive for judgment prediction. In contrast, PLJP, which uses fixed three-type reorganized case facts, may ignore some important details. FOL judgment rules help the model better understand the logical structure and legal terminology in complex facts, which can focus on important logical details to reduce incorrect judgments caused by excessively long texts.

\section{Conclusion}
In summary, our proposed RLJP framework introduces three key innovations for LJP: (1) A dynamic rule optimization method that formulates judgment rule optimization with CACL as the process of tree-splitting, effectively addressing fixed rules' adaptability limitations in complex legal cases; 
(2) RLJP is a logic-semantic co-reasoning architecture combining lightweight semantic prescreening with FOL judgment rules to strengthen judgment reasoning capabilities; 
(3) Experimental validation demonstrating superior performance over all baseline methods in LJP tasks, particularly in complex detailed cases.

\section*{Limitations}
Although our approach produces promising results on two public datasets, there are certain limitations. In the future, we will continue to dig into these concerns.

First, we only evaluate RLJP on two Chinese LJP datasets. We do not conduct experiments on other languages’ LJP datasets to evaluate the validity. Second, the interpretability of LJP is crucial, and users need to understand how the model can get the judgment prediction results. In the examination module, RLJP combined COT with FOL rules and output the explanation for the result, but it lacked sufficient interpretability analysis for the judgment process and results of the model.

\section *{Ethical Consideration}
While our RLJP framework enhances legal judgment prediction accuracy, its deployment requires rigorous ethical safeguards:
(1) Human judges must retain final decision-making authority to mitigate risks from data biases or model errors;
(2) Clear accountability protocols should ensure legal responsibility remains with human practitioners, not AI systems;
(3) Ongoing fairness audits are necessary to prevent discriminatory outcomes against demographic minorities.

\bibliography{custom}

\appendix

\section{The construction of quiz experience.}

\begin{figure}[hbtp] 
  \centering
  \includegraphics[width=5cm]{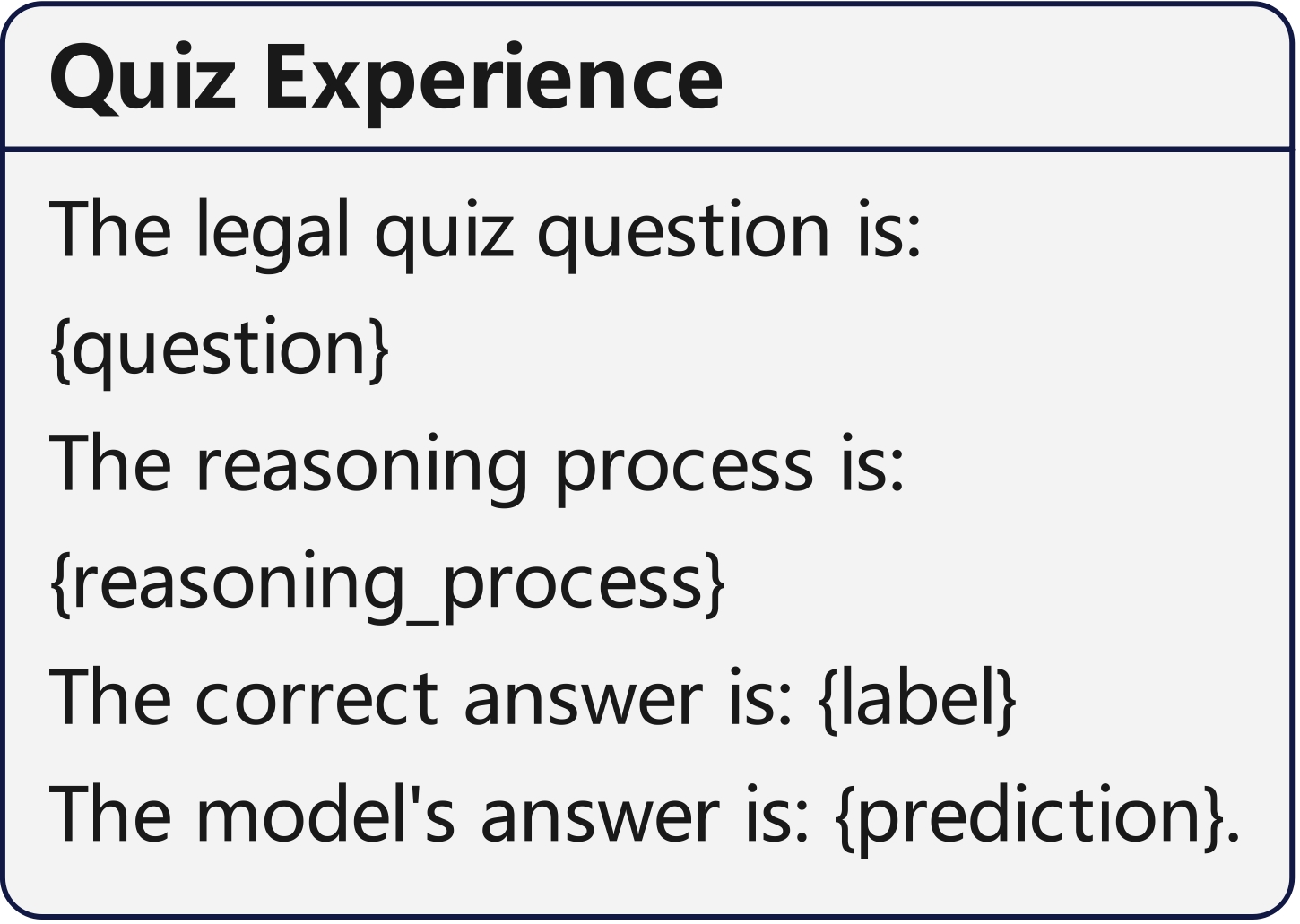} 
  \caption{The construction of quiz experience.}
  \label{fig: quiz_experience}
\end{figure}

\section{The prompt template for optimization direction generation.}
\begin{figure}[hbtp]
  \centering
  \includegraphics[width=5cm]{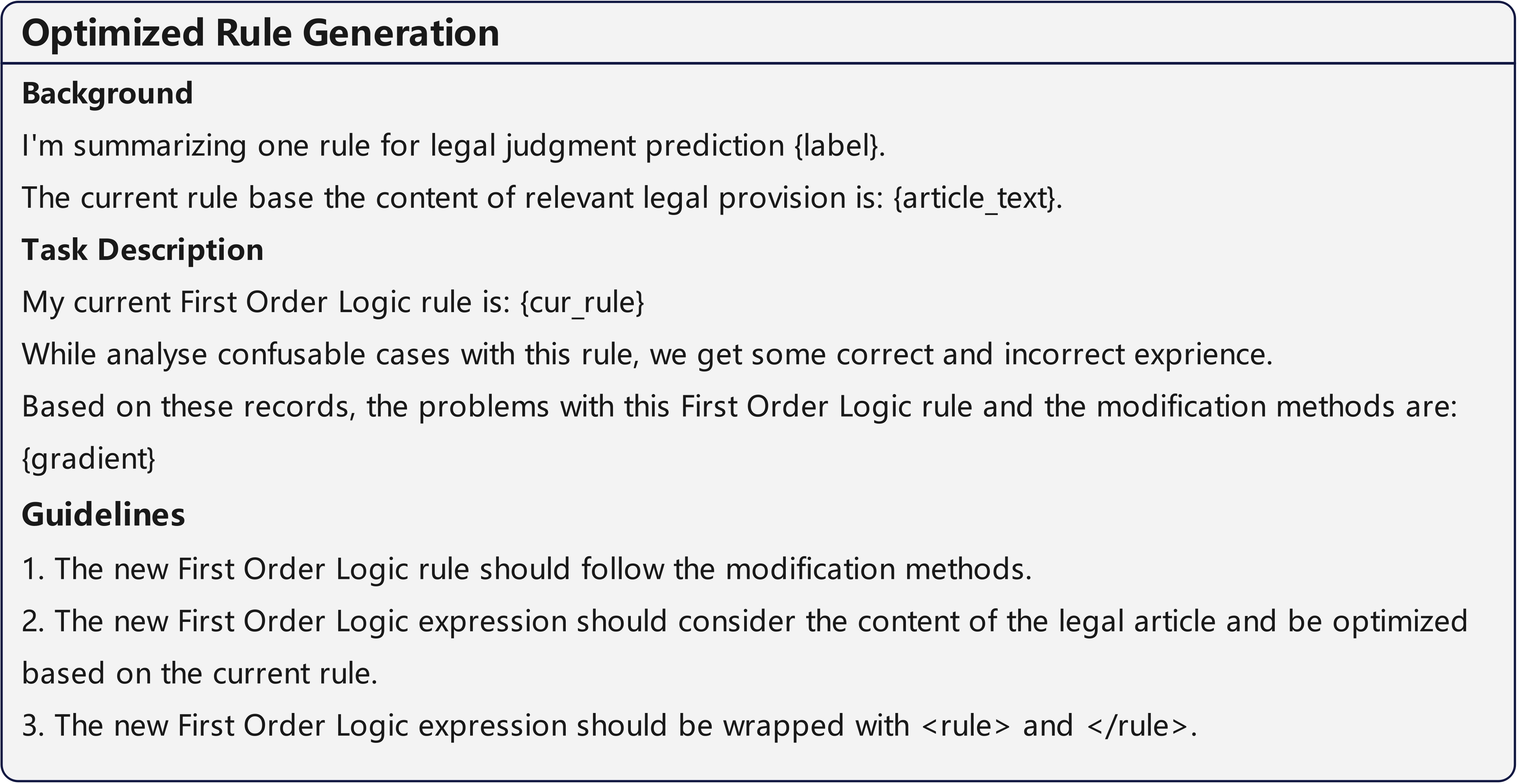} 
  \caption{The prompt template for optimization direction generation.}
  \label{fig: optRule}
\end{figure}

\section{The prompt template for the quiz in the confusable case set.}

\begin{figure}[hbtp]
  \centering
  \includegraphics[width=5cm]{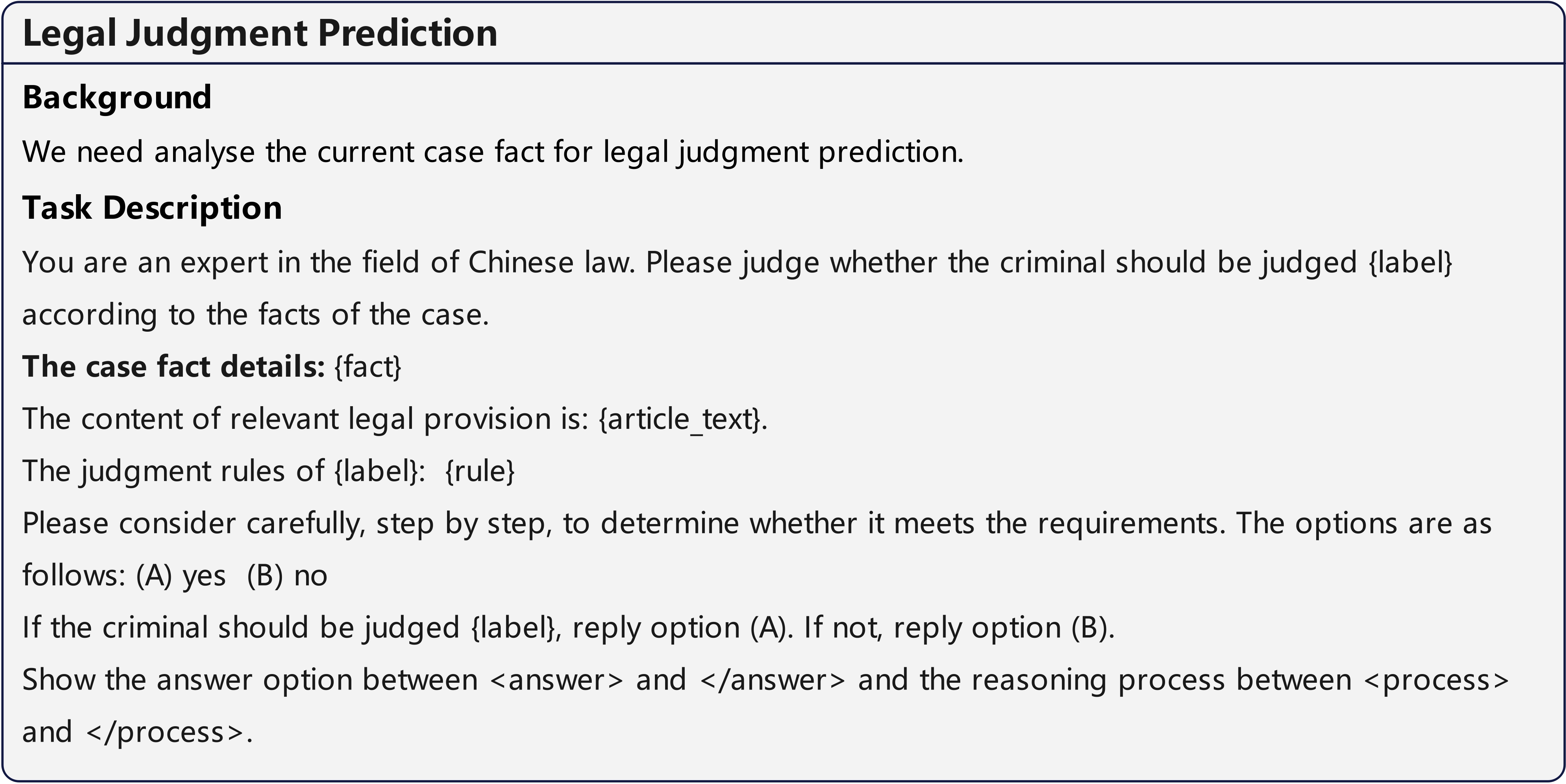} 
  \caption{The prompt template for the quiz in the confusable case set.} 
  \label{fig: ljp}
\end{figure}

\section{The prompt template for optimization direction generation.}
\begin{figure}[H]
  \centering
  \includegraphics[width=5cm]{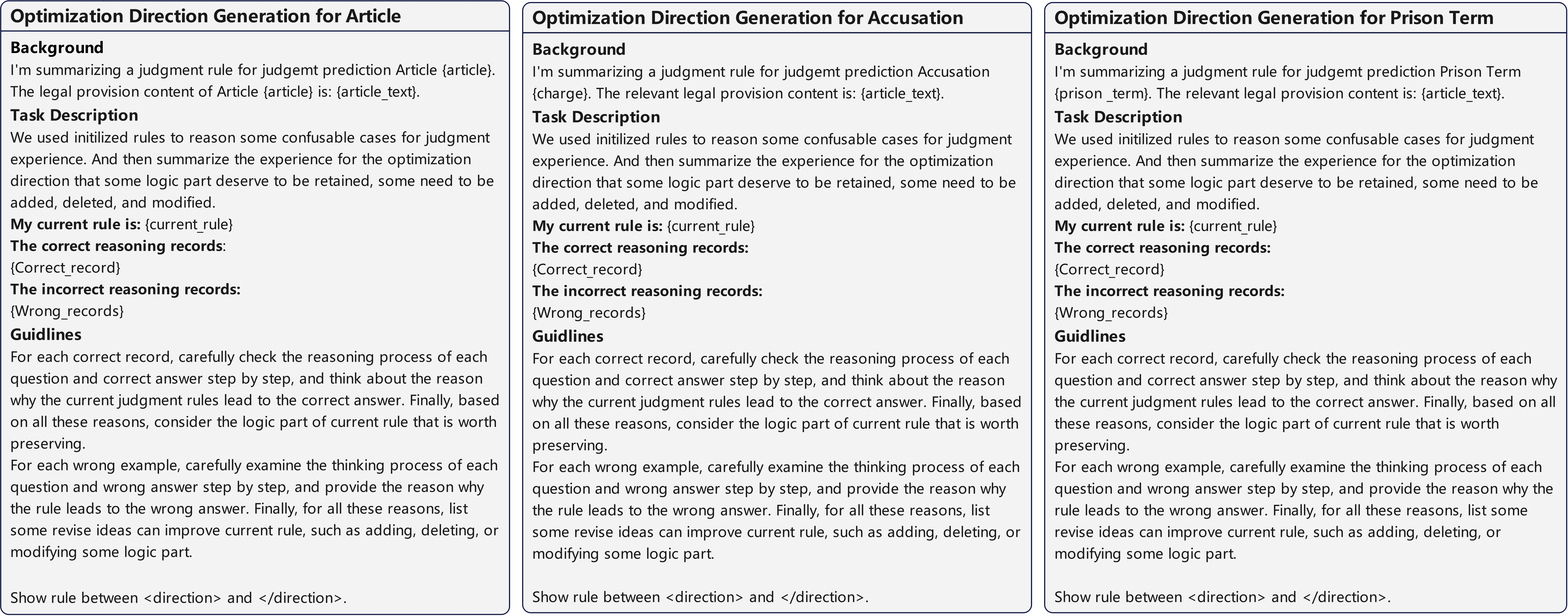} 
  \caption{The prompt template for optimization direction generation.}
  \label{fig: optDirection}
\end{figure}

\section{The prompt template for generating a case abstract.}

\begin{figure}[H]
  \centering
  \includegraphics[width=5cm]{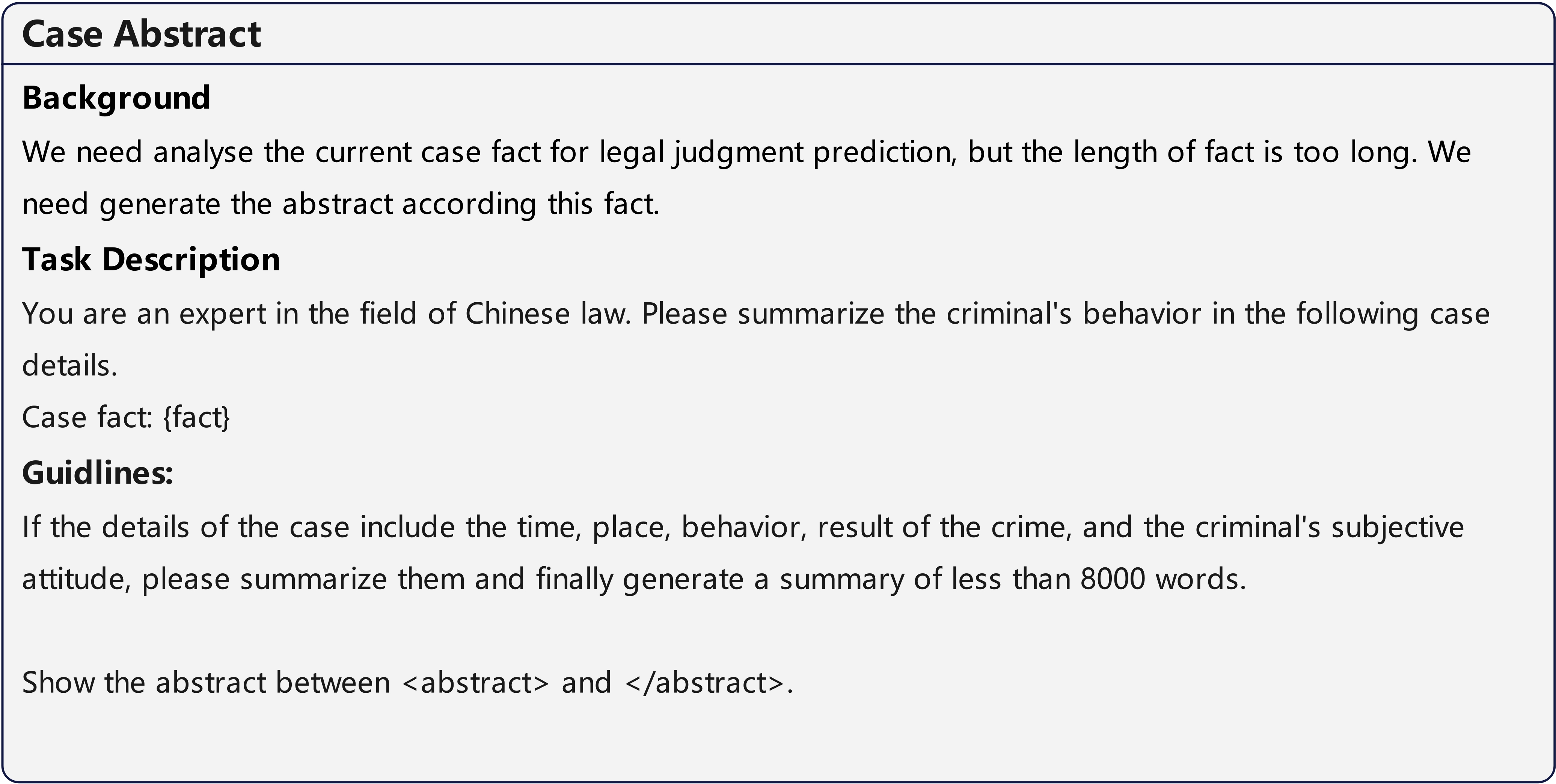} 
  \caption{The prompt template for generating a case abstract.}
  \label{fig: abstract}
\end{figure}

\section{The Statistics of LJP Datasets}
\label{sec:appendix_dataset}

Table \ref{tab: dataset} is the statistics of two realistic datasets, CAIL2018 and CJO22, which are widely used in the task of LJP. These datasets have been rigorously preprocessed to ensure compliance with ethical and legal standards. All personally identifying information (PII), including names, addresses, identification numbers, and other sensitive identifiers, has been anonymized or removed. Additionally, these datasets uncover offensive, discriminatory, or harmful content, as such material was systematically filtered during data curation. These measures align with established guidelines for ethical AI research and protect the privacy and dignity of individuals involved in legal cases.

\setlength{\tabcolsep}{1mm}{
\begin{table}[htbp]
\begin{tabular}{lcc}
\hline
\textbf{Type}          & \textbf{CAIL2018} & \textbf{CJO22} \\ \hline
Law Articles & 164               & 164            \\
Charges      & 42                & 42             \\
Prison Terms & 10                & 10             \\
Samples      & 82138             & 1698           \\ 
Avg. \# Fact words & 288.6             & 461.7  \\
\hline
\end{tabular}
\caption{Datasets statistics}
\label{tab: dataset}
\end{table}
}

\section{The Experimental setting}
\label{sec:appendix_set}
Our computing infrastructure was robust, featuring two A800 GPUs, each equipped with 80GB of memory, providing the necessary computational power to handle the large-scale data and complex algorithms involved in our experiments. Additionally, we utilized a high-performance computing cluster with a multi-core CPU architecture, high-speed NVMe storage for rapid data access, and a high-bandwidth network interconnect to ensure efficient data transfer and processing. The operating environment was Linux-based, and we employed CUDA and cuDNN libraries to optimize GPU performance. All these details are crucial for understanding the computational resources that underpinned our experimental results.

\section{The Statistics of Two Extracted Datasets}
\label{sec:appendix_long}

\setlength{\tabcolsep}{0.5mm}{
\begin{table}[h]
\begin{tabular}{lcc}
\hline
\textbf{Type}& \textbf{CAIL2018\_long} & \textbf{CJO22\_long} \\ \hline
Law Articles & 43  & 29            \\
Charges      & 41               & 28             \\
Prison Terms & 10                & 10             \\
Samples      & 7499             & 83           \\ 
Avg. \# Fact words & 1147.5             & 10815.9  \\
Max. \# Fact words & 20397   & 43030  \\
\hline
\end{tabular}
\caption{Test sets in the analytical experiment, which contain the top 5\% of cases in length from CAIL2018 and CJO22}
\label{tab:long_set}
\end{table}
}

Table \ref{tab:long_set} shows the two subsets statistics of two realistic datasets CAIL2018 and CJO22 as the test set used in the analytical experiment. The two subsets contain the top 5\% of cases sorted by case length from CAIL2018 and CJO22.

\end{document}